\documentclass[sigconf]{acmart}

\usepackage{booktabs} 

\usepackage{booktabs}
\usepackage{multirow}
\usepackage{multicol}
\usepackage{colortbl}
\usepackage{subfigure}

\usepackage{enumerate}

\begin{document}
\title{Image Quality Assessment Guided Deep Neural Networks Training}


\author{Zhuo Chen}
\affiliation{%
  \institution{Interdisciplinary Graduate School, Nanyang Technological University}
}
\email{ZCHEN036@e.ntu.edu.sg}

\author{Weisi Lin}
\affiliation{%
  \institution{School of Computer Science and Engineering, Nanyang Technological University}
}

\author{Shiqi Wang}
\affiliation{%
  \institution{Department of Computer Science, City University of Hong Kong}
}

\author{Long Xu}
\affiliation{%
  \institution{Key Laboratory of Solar Activity, National Astronomical Observatories, Chinese Academy of Sciences}
}

\author{Leida Li}
\affiliation{%
  \institution{School of Information and Control Engineering, China University of Mining and Technology}
}


\begin{abstract}
For many computer vision problems, the deep neural networks are trained and validated based on the assumption that the input images are pristine (i.e., artifact-free). However, digital images are subject to a wide range of distortions in real application scenarios, while the practical issues regarding image quality in high level visual information understanding have been largely ignored. In this paper, in view of the fact that most widely deployed deep learning models are susceptible to various image distortions, the distorted images are involved for data augmentation in the deep neural network training process to learn a reliable model for practical applications. In particular, an image quality assessment based label smoothing method, which aims at regularizing the label distribution of training images, is further proposed to tune the objective functions in learning the neural network. Experimental results show that the proposed method is effective in dealing with both low and high quality images in the typical image classification task.
\end{abstract}


\maketitle

\section{Introduction}
\label{sec:intro}

Recently, deep neural networks (DNNs) have demonstrated state-of-the-art performance in various computer vision tasks, e.g., face recognition \cite{sun2014deep,taigman2014deepface}, pedestrian detection \cite{ouyang2013joint} and pose estimation \cite{toshev2014deeppose}. In contrast to the handcrafted features such as Scale-Invariant Feature Transform (SIFT) \cite{lowe2004distinctive}, deep learning based approaches are able to learn representative features directly from the vast amounts of data. For general nature image classification, which is of great interest to DNN models, the AlexNet model \cite{krizhevsky2012imagenet} has achieved 9\% better classification accuracy than the previous hand-crafted methods in the 2012 ImageNet competition \cite{russakovsky2015imagenet}, which provides a large scale training dataset with 1.2 million images and one thousand categories. Subsequently, inspired by such fantastic progress, DNN models continue to be the undisputed leaders in the competition of ImageNet. In particular, both VGGNet \cite{simonyan2014very} and GoogLeNet \cite{szegedy2015going} announced promising performance in the ILSVRC 2014 classification challenge, which demonstrated that deeper and wider architectures can bring great benefits in learning better representations via large scale datasets. For face recognition, DeepID \cite{sun2014deep} trained on 202,599 face images of 10,177 identities has achieved 96.05\% accuracy on LFW, and DeepFace \cite{taigman2014deepface} developed by Facebook yielded 97.35\% accuracy with 4.4 million faces of 4,030 identities data. Moreover, Google \cite{schroff2015facenet} used over 100 million face samples from 8 million identities to train DNN models, which achieved 99.63\% accuracy on LFW.

\begin{figure*}
	\centering
	\includegraphics[width=145mm]{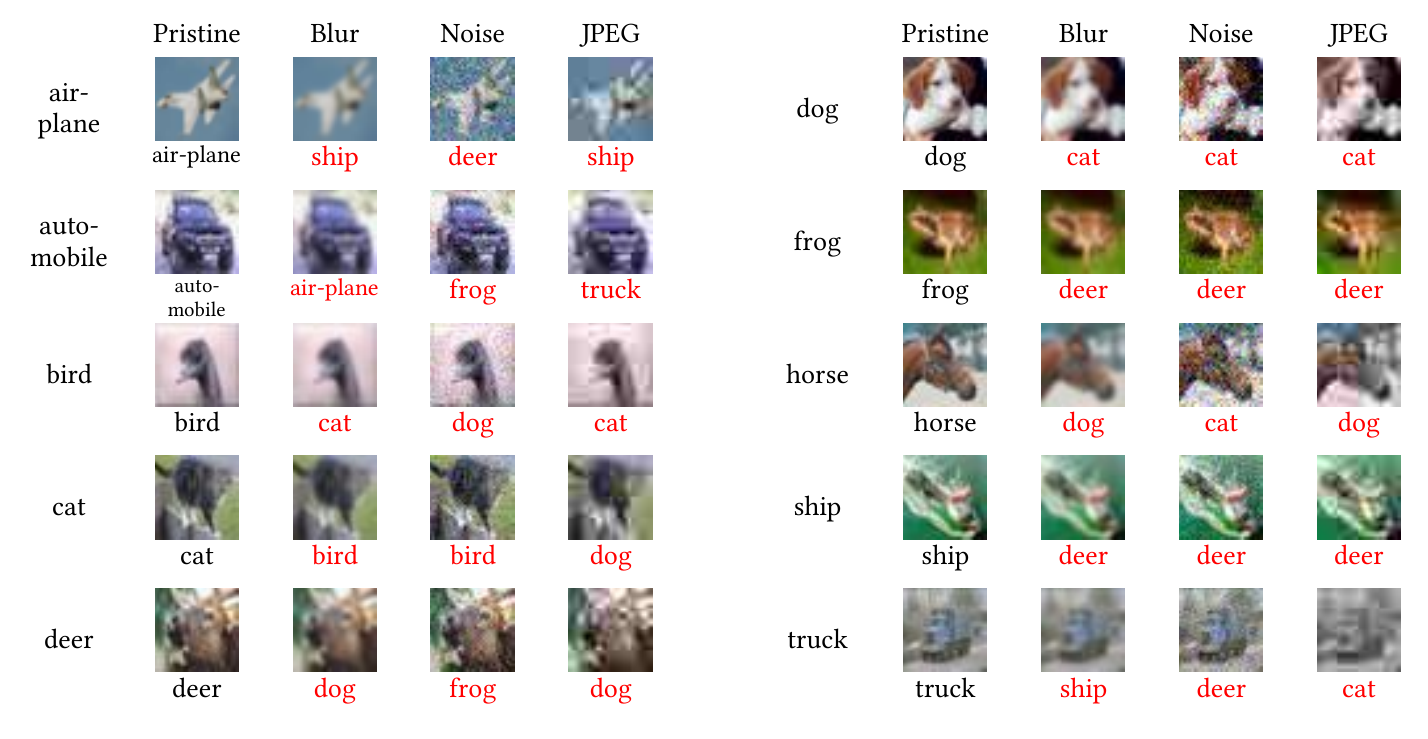}
	\vspace{-5 mm}
	\caption{Illustration of the classification results of the 10 classes in CIFAR-10, where each class set including one pristine sample image and its corresponding distorted images. The first column of a class set is the pristine sample image of CIFAR-10. The second, third and fourth columns are the images corrupted by blur, noise and JPEG compression respectively. Given the pristine images, a DNN model can correctly identify them. However, when distortions are injected, the DNN model may misclassify the inputs. The labels below thumbnails are the prediction results by the DNN model.}
	\label{fig:samples}
\end{figure*}

Generally speaking, DNN is a data driven method making it feasible to achieve outstanding performance with the explosion of big data. However, such property gives rise to the fact that the capability of deep models heavily relies on the training samples. In particular, most DNN models were trained and tested based on the assumption that the input image samples are pristine without any distortions injected. As such, they can achieve promising performance on high quality samples, but the performance will be seriously degraded when encountering with low quality images. Fig.~\ref{fig:samples} provides some examples in CIFAR-10 dataset \cite{krizhevsky2009learning} and it is shown that DNN model fails in predicting the correct classes when the input images are distorted. A recent work \cite{dodge2016understanding} evaluated several classical deep models for image classification by injecting different types of distortion into the test images. The results show that all the evaluated neural networks are susceptible to typical distortions such as blur and noise. For example, more than 20\% Top 1 and Top 5 accuracy drop can be observed when the images are distorted by Gaussian blur.

In real application scenarios, distortions will be introduced in image acquisition, compression, processing, transmission and reproduction. Generally speaking, restoration of such distorted images is an ill-posed problem, and even state-of-the-art algorithms cannot efficiently remove such artifacts. Therefore, evaluating the visual quality of these distorted images becomes meaningful. In the literature, there are numerous approaches proposed to assess the degradation of visual quality \cite{lin2011perceptual}. Popular image quality assessment (IQA) algorithms such as SSIM \cite{wang2004image}, FSIM \cite{zhang2011fsim}, GSIM \cite{liu2012image}, VSNR \cite{chandler2007vsnr}, PCQI \cite{wang2015patch}, etc., focus on the perception of quality degradation from the perspective of viewing experience. Due to the fact that the distortions can also bring difficulties in image understanding, it becomes more and more important to further investigate the applications of these IQA algorithms in the context of computer vision, as computer vision systems aim to automatically achieve the high-level understanding tasks that the human visual system can perform.

This naturally inspires us to incorporate the quality measure in the DNN learning process to deal with the visual understanding with low quality images. In particular, we first train the deep neural network by augmenting the data with the mixture of pristine and distorted data. Then an IQA-based label smoothing technique is proposed to enhance the performance of deep models by fine-tuning the network with the IQA measure. In this manner, the robustness of DNN models with distorted input data can be significantly improved. Experimental results show that the proposed scheme can significantly improve the classification performance of both high and low quality images.

The main contributions of the paper are as follows:

\begin{enumerate}[1)]
	\item We investigate in training deep models with data augmentation by both high and low quality images, and in-depth analyses of involving low quality images in the training process are also provided.
	\item We make a further attempt to adopt the IQA measure for label smoothing in the deep neural network training process. This leads us to the robust neural network that provides promising prediction performance for images with a broad range of quality levels.
\end{enumerate}

The rest of the paper is organized as follows. In Section~\ref{sec:dist_type}, we analyze the data augmentation with distorted images for deep learning. In Section~\ref{sec:iqals}, the proposed scheme with IQA-based label smoothing is introduced. Section~\ref{sec:experiment} provides the experimental results and analyses. Finally, Section~\ref{sec:conclusion} concludes this paper.

\begin{figure*}
	\centering
	\includegraphics[width=120mm]{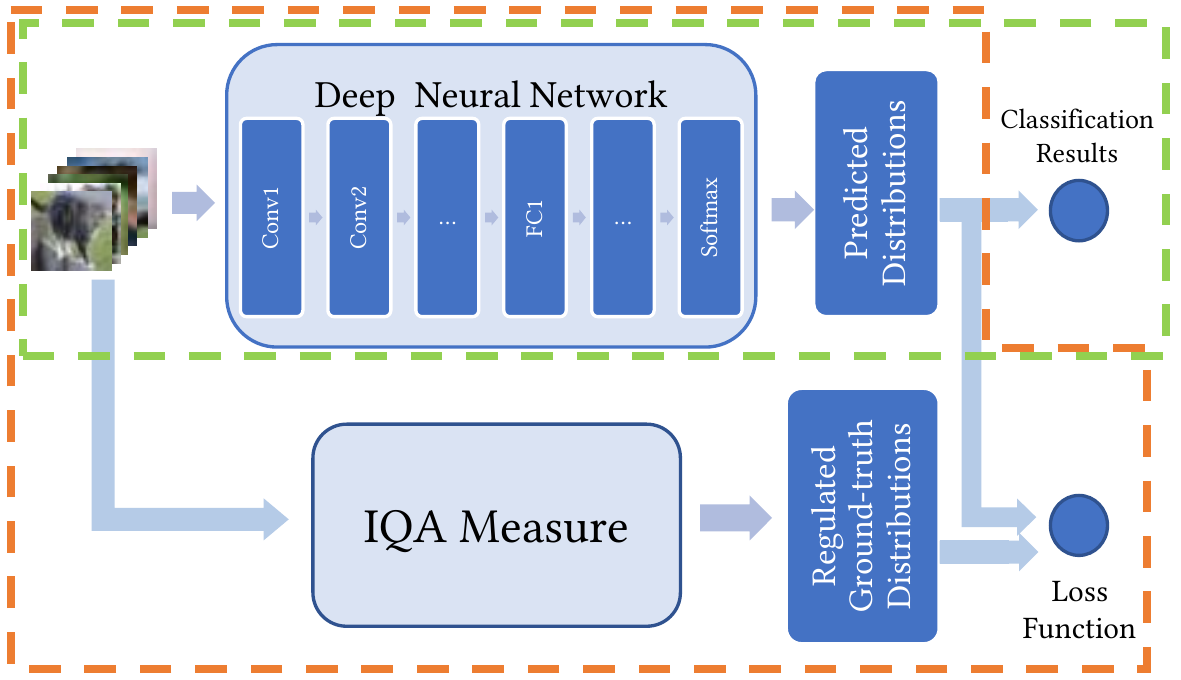}
	\caption{The deep learning framework with IQA-based label smoothing. The modules in the orange dashed box represent the training process while the modules in the green dashed box stand for the inference process.}
	\label{fig:framework}
\end{figure*}

\section{Data Augmentation with Distorted Images for Deep Learning}
\label{sec:dist_type}
It is generally acknowledged that the capability of the deep model largely depends on the training data. As such, to deal with the low quality test images, the straightforward solution is augmenting the training data with low quality versions and mixing them together with the original high quality data. In view of this, in this work we involve the pristine as well as the distorted images as the input data to learn the CNN model. The distorted versions are generated by injecting different levels of distortion into the pristine training images manually. As one of the first attempts using distorted images as the training data, here we are particularly interested in three types of commonly encountered distortions: blur, noise and JPEG compression.

Blurring artifacts are introduced due to the limitations of acquisition process such as camera motion or out-of-focus of the optical system. In addition, necessary image processing operations may also create blurring artifacts, such as denoising and compression. Therefore, investigating the blurring artifacts enables us to simulate the real-world application scenarios in recognizing and understanding the blurry images. Blurring artifacts will dramatically reduce the sharpness of images and distort the structural information of the texture content, which directly influence the interpretation of the images.

Noise may result from acquiring images with low quality camera sensors or in poor illumination conditions. Generally speaking, Gaussian noise can be used to model this scenario by adding the disturbance to each pixel. The introduced noise will make the image more disorderly with higher uncertainties for human perception from the perspective of the free energy theory \cite{zhai2012psychovisual}.

Lossy compression techniques are also commonly used in computer vision processing, which introduce distortion in visual signals while reducing storage or memory footprint requirement simultaneously. Among the digital image coding standards, JPEG compression is the most wildly-used one, which brings strong blocking boundaries and ringing artifacts in low bit coding scenarios.

All these types of distortions may introduce disturbance in representing the semantic information of images by damaging the image content. According to the study in \cite{dodge2016understanding}, not only the human visual system, but also computer vision algorithms cannot efficiently recognize and understand the visual information with these three kinds of distortions. Therefore, it is meaningful to start from these artifacts, which pose a unique set of challenges to computer vision tasks.
However, there are also side effects when both pristine and distorted images are used as the training data. In particular, although the model can provide promising prediction performance on the distorted images in principle, it may also significantly degrade the testing performance when we feed the high quality images as the input. Thus, in this work we further seek a good balance between the high quality and low quality training images with IQA to learn a more robust model.

\section{IQA-based Label Smoothing}
\label{sec:iqals}
In typical image classification tasks, a softmax layer is usually laid on the top of a neural network to predict the probabilities of an input image belonging to each given class. Alternatively, the probability can also be treated as the perceptive confidence of the model for the input image belonging to a specific class. For each input image $x$, the softmax layer computes the probability $p(k|x)$ for each given label $k\in \{ 1\cdots K \}$. The ground-truth distribution is denoted by $q(k|x)$. For brevity, we omit the dependence of $p$ and $q$ on the input $x$. During the training process, the loss can be calculated with $p$ and $q$ as follows,
\begin{equation}
l=-\sum_{k=1}^{k}(p(k)-q(k))^2.
\end{equation}
As such, minimizing the loss is equivalent to maximizing the expected Euclidean-likelihood of a label, where the label is selected according to its ground-truth distribution $q(k)$. In most situations, given a single ground-truth label $y$, $q(y)=1$ for $k=y$ while $q(k)=0$ when $k\neq y$. This encourages the deep model to be robust and confident in the classification tasks of pristine images.

As mentioned early, although the corrupted images are involved in the training process to deal with the scenario of low quality images as the input, such method may not permanently solve the problem, and there are several challenging issues:
\begin{enumerate}[1)]
	\item Since both high and low quality images are used as the training data, the learned model is lack of the generalization capability and exhibit strong bias to the low quality images. As such, though we can improve the accuracy of low quality images, the prediction performance of the high quality images will be degraded.
	\item The human visual perception has been largely ignored in the learning process. It is generally hypothesized that the human visual system evolves through learning from the natural images that possess certain statistical properties. As such, low quality images which belong to unnatural images should play a less importance role compared with pristine images since low-quality images are more difficult to understand. As such, a reasonable way to manipulate this is to make the expected probabilities corresponding to the ground-truth labels of the low-quality images lower. However, the commonly used ground-truth distribution does not follow this trend.
\end{enumerate}

To avoid these drawbacks and incorporate the brain-like perception in the deep learning framework, an IQA-based label smoothing method (IQA-LS) is proposed. In particular, given the label $k$ and a single input $x$ with ground-truth label $y$, the label distribution $q(k|x)$ is reformulated as follows
\begin{equation}
q{'}(k|x)=\begin{cases}
T(s(x)) &  k= y \\
(1-T(s(x)))/(K-1) &   k\neq y
\end{cases}
\end{equation}
where $s(\cdot )$ denotes the score of IQA measure and $T(\cdot )$ transforms the IQA score to the range of $(0,1]$. This implies that the distribution of the label $k$ is obtained based on the IQA score of the input image $x$ when $k= y$, while the uniform distribution is employed for the rest labels. Therefore, the confidence value is directly determined by the image quality, and better quality implies higher confidence in the network learning. This is in line with the human perception when understanding the image content, as low quality images may be perceived with higher uncertainties from the perspective of free-energy theory \cite{zhai2012psychovisual}.

In this work, we adopt the SSIM \cite{wang2004image} as the IQA measure and identity function (i.e. $f(x)=x$) for $T(\cdot )$, due to its good trade-off between the accuracy and computational complexity. In particular, it is computed by comparing the original and distorted images based on the degradation of the structural information. IQA measure is only employed in the training procedure where we can get access to both the distorted and its corresponding pristine images, as shown in Figure~\ref{fig:framework}. It is also worth mentioning that other IQA algorithms including reduced-reference (RR) and no-reference (NR) methods are also compatible with our proposed IQA-based label smoothing framework.

\section{Experimental Results}
\label{sec:experiment}

To evaluate the reliability of DNN models learned by the proposed IQA-based label smoothing method, experiments are conducted on the CIFAR-10 database. First, we briefly introduce the database and the training data enhancement strategy. Subsequently, the parameter settings in the experiments are detailed. Finally, we illustrate and analyse the classification results with different training strategies.

\subsection{Dataset description}
In this paper, the proposed scheme is evaluated on CIFAR-10 dataset which is a labelled subset of 80 million tiny images \cite{torralba200880}. As shown in Figure~\ref{fig:samples}, CIFAR-10 is composed of colour images with size $32\times 32$, and in total there are 10 different classes for performing the classification tasks. Moreover, the dataset contains 50,000 training samples and 10,000 testing samples. In order to compare the models trained with different strategies, except for specific image distortion, data enhancement strategies (e.g. image contrast, brightness and saturation adjustment) are prohibited in the training process. Here, as discussed in Section~\ref{sec:dist_type}, we only apply Gaussian blur, Gaussian noise and JPEG compression on the pristine training images.

\subsection{Benchmark learning architecture}
\label{sec:arch}
The learning architecture is designed following the model described in the Tensorflow \cite{abadi2016tensorflow}. In particular, it is a Tensorflow based duplication of Alex Krizhevsky's work \cite{krizhevsky2012imagenet} with a few modifications. Specific descriptions regarding the proposed learning architecture is illustrated in Table~\ref{tab:arch}.
\begin{table}
	\centering
	\caption{Descriptions of the learning architecture. The input and output sizes are specified as $rows\times cols\times channels$, and the kernel is characterized in terms of $rows\times cols ,stride$.}
	\label{tab:arch}
	\begin{tabular}{cccc}
	\toprule
		layer & size-in & size-out & kernel\\
		\midrule
		$conv1$ & $32\times 32\times 3$ & $32\times 32\times 64$ & $5\times 5, 1$\\
		$pool1$ & $32\times 32\times 64$ & $16\times 16\times 64$ & $3\times 3, 2$\\
		$lrn1$ & $16\times 16\times 64$ & $16\times 16\times 64$ & / \\
		$conv2$ & $16\times 16\times 64$ & $16\times 16\times 64$ & $5\times 5, 1$\\
		$lrn2$ & $16\times 16\times 64$ & $16\times 16\times 64$ & / \\
		$pool2$ & $16\times 16\times 64$ & $8\times 8\times 64$ & $3\times 3, 2$\\
		$fc1$ & 4096 & 384 & / \\
		$fc1$ & 384 & 192 & / \\
		$softmax$ & 192 & 10 & / \\
	\bottomrule
\end{tabular}
\end{table}

This model follows the common multi-layer convolutional neural network (CNN) architecture which consists of alternating convolutions and nonlinearities. This architecture takes the full-size image as the input. The padded input is filtered with 64 kernels of size $5\times 5\times 3$ with a stride of 1 pixel to produce 64 feature maps with identical size as the input. After the nonlinear transformation by Rectified Linear Units (ReLU), the feature maps are response-normalized and down-sampled by Local Response Normalization (LRN) and Overlapping max-Pooling respectively \cite{krizhevsky2012imagenet} to generate the input of second convolutional layer. Subsequently, the similar process is applied again to produce the input of the following fully connected layers whose activation function is also ReLU. Finally, the softmax layer takes the representation learned by the network to create the final classification results, which characterize the probabilities of the input belonging to each given class.

\begin{table*}[t]
	\centering
	\caption{Performance comparisons of the models with different training strategies.}
	\label{tab:results}
	\begin{tabular}{ccc|cccccccccc}
	\toprule[0.3mm]
	\multirow{2}{*}{Strategy} & Regularization & Training & \multirow{2}{*}{Pristine} & \multicolumn{3}{c}{Blur}    & \multicolumn{3}{c}{Noise}   & \multicolumn{3}{c}{JPGE}    \\ \cline{5-13}
	    & Method     & Set        &                           & Level 1 & Level 2 & Level 3 & Level 1 & Level 2 & Level 3 & Level 1 & Level 2 & Level 3 \\ \midrule[0.3mm]
	1   & Original   & $Pristine$         & 0.794                     & 0.676   & 0.524   & 0.436   & 0.677   & 0.566   & 0.392   & 0.600   & 0.529   & 0.391   \\
	2   & Original   & $MIX_{blur}$       & 0.781                     & \cellcolor[gray]{0.8} 0.777   & \cellcolor[gray]{0.8} 0.766   & \cellcolor[gray]{0.8} \textbf{0.751}   & 0.735   & 0.681   & 0.585   & 0.669   & 0.621   & 0.469   \\
	3   & IQA-LS     & $MIX_{blur}$       & \textbf{0.798}            & \cellcolor[gray]{0.8} \textbf{0.782}   & \cellcolor[gray]{0.8} \textbf{0.766}   & \cellcolor[gray]{0.8} 0.749   & 0.749   & 0.698   & 0.607   & 0.681   & 0.624   & 0.465   \\
	4   & Original   & $MIX_{noise}$      & 0.794                     & 0.716   & 0.606   & 0.526   & \cellcolor[gray]{0.8} 0.779   & \cellcolor[gray]{0.8} 0.773   & \cellcolor[gray]{0.8} \textbf{0.749}   & 0.693   & 0.640   & 0.496   \\
	5   & IQA-LS     & $MIX_{noise}$      & \textbf{0.807}            & 0.736   & 0.612   & 0.525   & \cellcolor[gray]{0.8} \textbf{0.792}   & \cellcolor[gray]{0.8} \textbf{0.776}   & \cellcolor[gray]{0.8} 0.743   & 0.699   & 0.642   & 0.503   \\
	6   & Original   & $MIX_{JPEG}$       & 0.753                     & 0.728   & 0.711   & 0.656   & 0.720   & 0.662   & 0.616   & \cellcolor[gray]{0.8} 0.710   & \cellcolor[gray]{0.8} 0.687   & \cellcolor[gray]{0.8} \textbf{0.607}   \\
	7   & IQA-LS     & $MIX_{JPEG}$       & \textbf{0.791}            & 0.751   & 0.704   & 0.622   & 0.742   & 0.667   & 0.607   & \cellcolor[gray]{0.8} \textbf{0.722}   & \cellcolor[gray]{0.8} \textbf{0.693}   & \cellcolor[gray]{0.8} 0.590   \\ \hline
	8   & Original   & $MIX_{all3}$       & 0.767                     & 0.753   & 0.737   & 0.721   & 0.761   & 0.744   & 0.725   & 0.721   & 0.686   & \textbf{0.599}   \\
	9   & IQA-LS     & $MIX_{all3}$       & \textbf{0.790}            & \textbf{0.773}   & \textbf{0.753}   & \textbf{0.738}   & \textbf{0.771}   & \textbf{0.757}   & \textbf{0.731}   & \textbf{0.726}   & \textbf{0.692}   & 0.585   \\
	\bottomrule[0.3mm]
	\end{tabular}
\end{table*}

\subsection{Parameter setting}
The proposed deep architectures are trained with stochastic gradient descent method on a NVIDIA GeForce 980Ti GPU with batch size 100 for 2,000 epochs. All our experiments use the initial learning rate of 0.1 which decays for every 350 epochs with an exponential rate of 0.1. In addition, L2Loss weight decay multiplied by 0.004 is added to the two full-connected layers.

Regarding the training data, pristine images and the mixture data with both pristine and distorted images are used for validations. The pristine data are totally from the CIFAR-10 training dataset, and the mixture data are the combination of pristine and distorted images with different distortion levels in a fixed ratio. In each training epoch, 60\% of the pristine training samples are maintained, 15\% of them are with $level\ 1$ distortion, another 15\% of them are with $level\ 2$ distortion, and the rest 10\% samples are distorted in $level\ 3$.

Specifically, for the blur distortion, we use the Gaussian kernels with $\sigma = 0.7, 1.0, 1.2$ for $levels$ from 1 to 3 respectively. With respect to the noise artifacts, three levels of white Gaussian noise with variance values $v = 0.005, 0.01, 0.02$ are employed. Regarding to JPEG compression, we compress the images with the JPEG quality factors of $12, 8, 4$ corresponding to the distortion $levels$ from 1 to 3. In particular, the images generated by the selected parameters can severely influence the inference performance of deep networks, which has also been pointed out in \cite{dodge2016understanding}.

In summary, five different types of training set are utilized in the experiments: one pristine training set and four sets of mixture data with both pristine and distorted samples. For convenience, the four sets of mixture data are denoted as $MIX_{blur}$, $MIX_{noise}$, $MIX_{JPEG}$ and $MIX_{all3}$, where $MIX_{blur}$, $MIX_{noise}$ and $MIX_{JPEG}$ represent the training sets of mixture data with pristine and distorted images degraded by one certain type of distortion (blur, noise and JPEG respectively), while $MIX_{all3}$ is the combination of pristine and all the three types of distorted samples.

With respect to the testing data, a pristine set and nine distorted sets (with three different types and each one has three distortion levels) are generated for evaluating each learned model. Example images with the three types of artifact are also illustrated in Fig.~\ref{fig:samples}.

\subsection{Performance comparisons}
Nine different training approaches are implemented to evaluate the performance of the proposed scheme. All these nine approaches share the same architecture, as introduced in the sub-section~\ref{sec:arch}, but different training strategies. These training strategies are with different combinations of data augmentation strategies and label distributions. Here, we will detail these training strategies and analyse their performance in terms of the classification accuracy, as illustrated in Table~\ref{tab:results}.

\subsubsection{Training on pristine dataset}

Strategy 1 in Table~\ref{tab:results} aims to train deep models with the pristine CIFAR-10 training samples without the augmented data. Moreover, the label distribution of each training image is the classical 0-1 distribution. Such training strategy follows the widely adopted benchmark models such as AlexNet and VGG. Therefore, we consider this approach as the baseline.

From Table~\ref{tab:results}, we can see that the model trained with this strategy is sensitive to all the three involved distortions as the classification performance decreases dramatically when the distortion level increases. For instance, even with the blur $level$ 1 where the $\sigma$ value of Gaussian kernel is moderate, the accuracy drops more than 10\%. Similar trends can be observed for noise and JPEG compression artifacts. This is in accordance with the evaluation results in \cite{dodge2016understanding}. This phenomenon can be explained by the fact that the distortions can heavily remove the texture and edge information in an image, which is important to the DNN models learned with pristine images since such DNN models may always attempt to look for specific textures and edges for the classification task.

\subsubsection{Training on mixture dataset}

Strategies 2,4,6,8 in Table~\ref{tab:results} train DNN models with the mixture data of pristine and distorted images while the label distribution maintains the typical 0-1 distribution. Such kind of training approach is a straightforward solution to make the network better adapt to the distorted images.

\begin{figure*}
\centering
\subfigure[\small{}]
{\includegraphics[width=6.5cm]{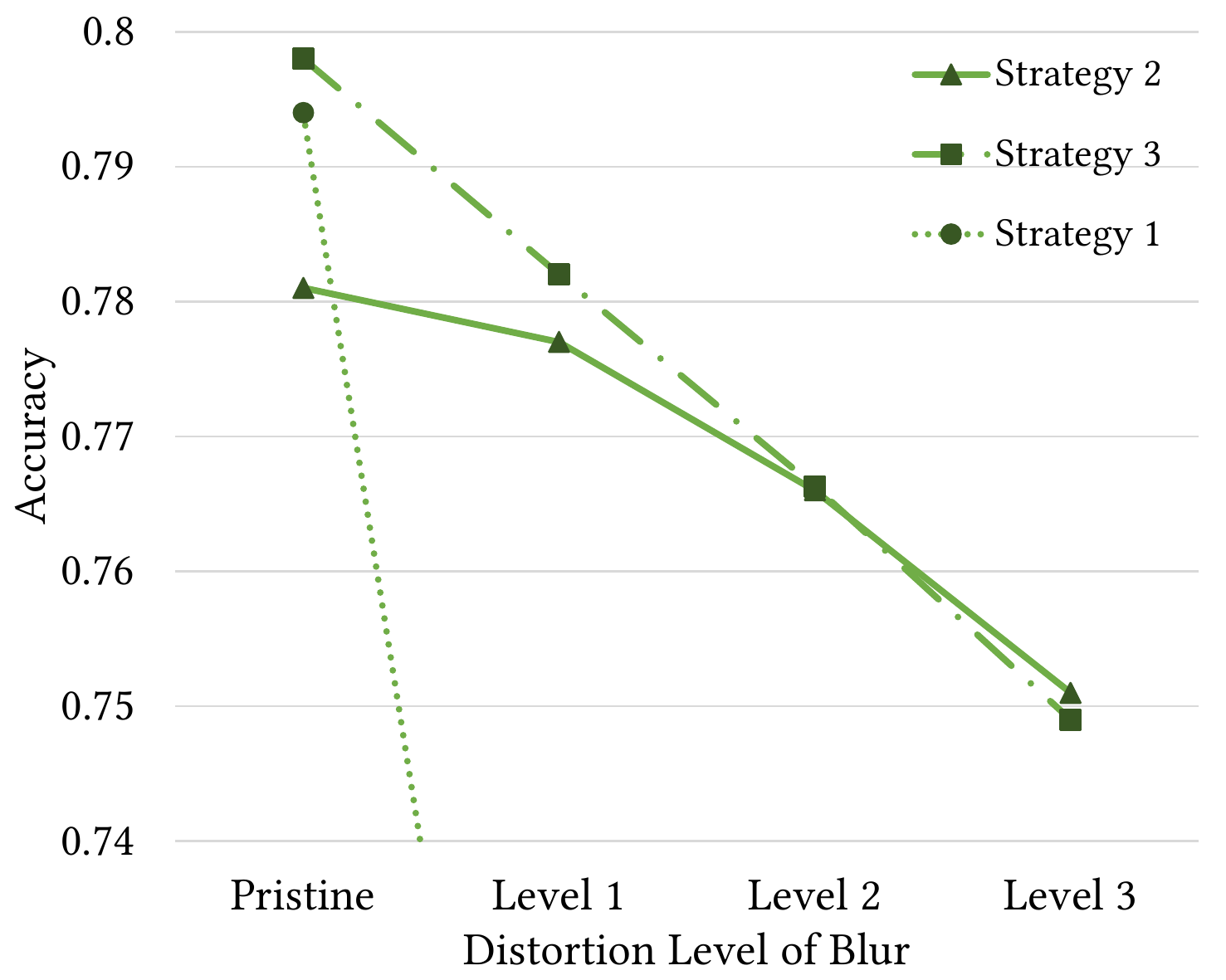}}\label{fig:res_comp_a}
\hspace{5mm}
\subfigure[\small{}]
{\includegraphics[width=6.5cm]{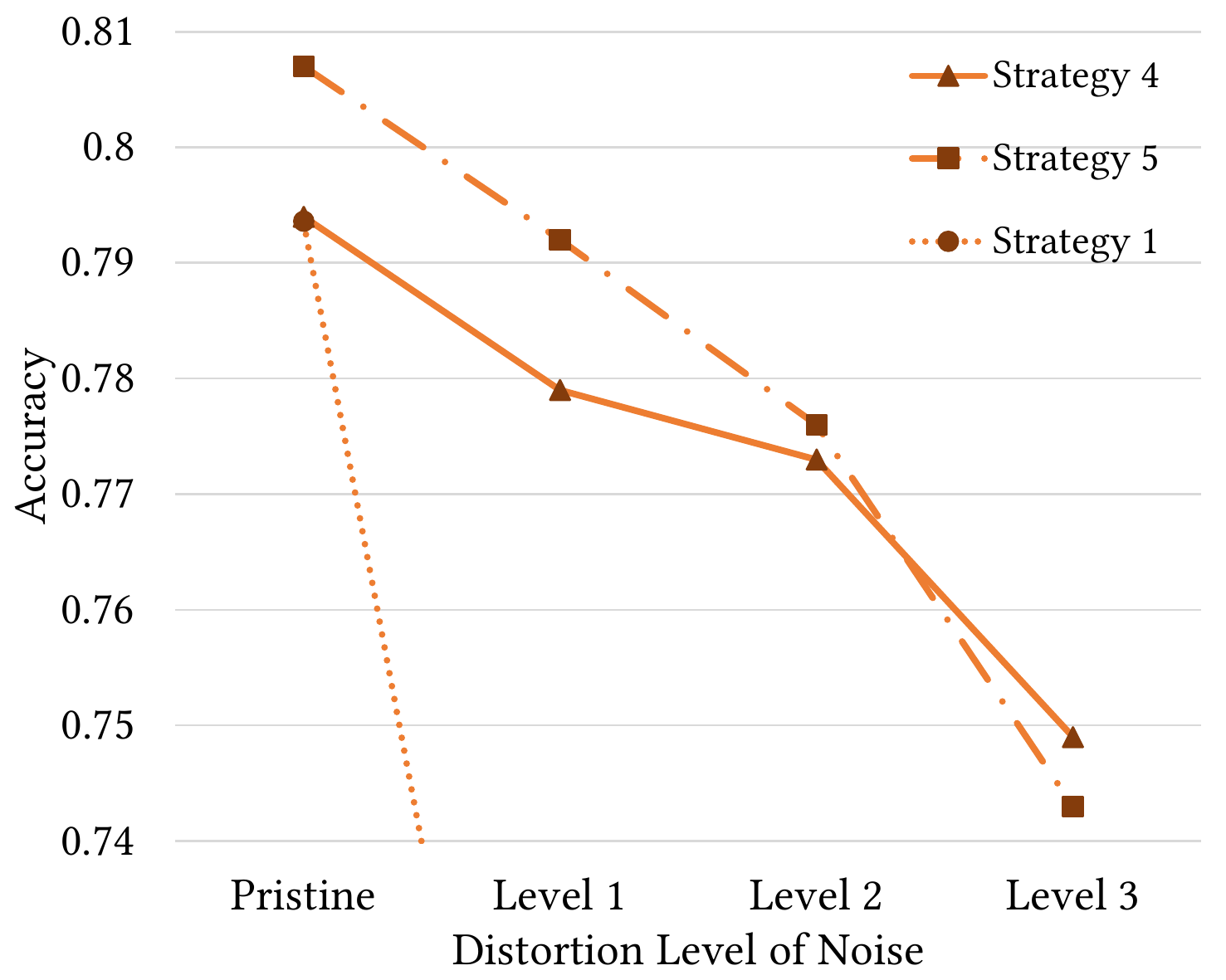}}\label{fig:res_comp_b}
\subfigure[\small{}]
{\includegraphics[width=6.5cm]{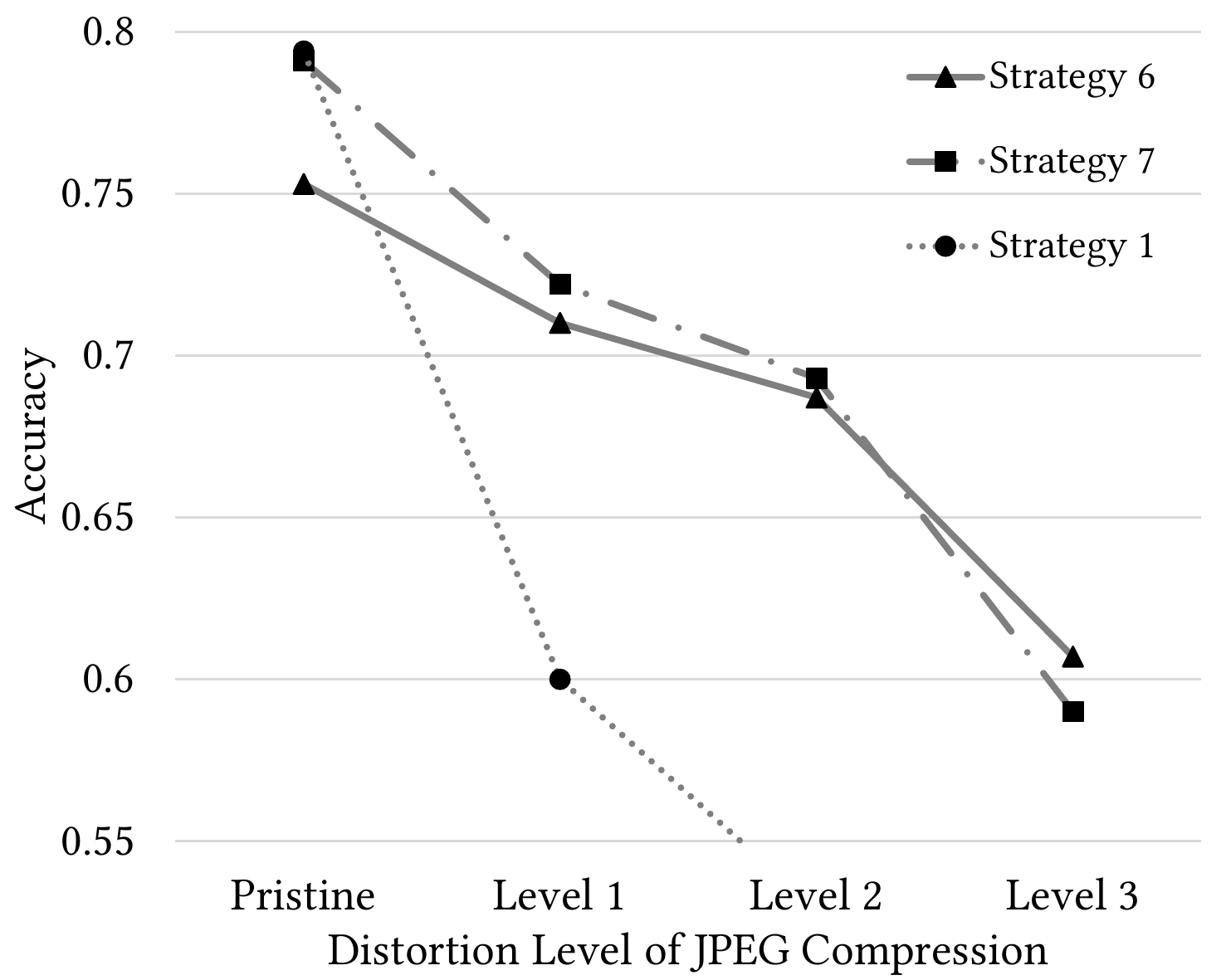}}\label{fig:res_comp_c}
\hspace{5mm}
\subfigure[\small{}]
{\includegraphics[width=6.5cm]{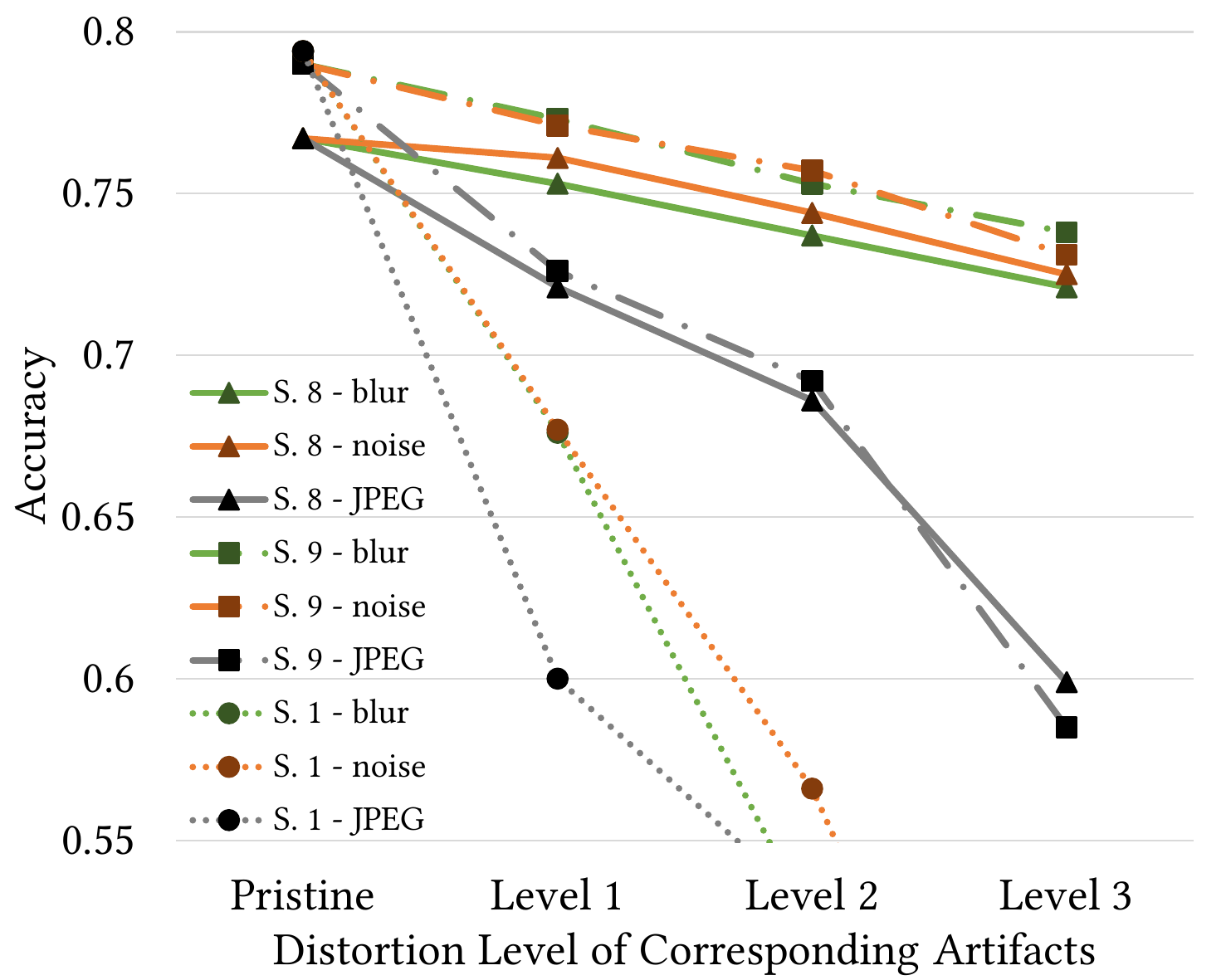}}\label{fig:res_comp_d}
\vspace{-3 mm}
\caption{Performance comparisions. (a) Performance comparisons of the models trained and tested with the distortion type of blur, which correspond to Strategies 2 and 3; (b) Performance comparisons of the models trained and tested on the distortion type of noise, which correspond to Strategy 4 and 5; (c) Performance comparisons of the models trained and tested on the distortion type of JPEG, which correspond to Strategy 6 and 7; and subfigure (d) Performance comparisons of the models trained on all three types of distorted images, and tested for each distortion type individually (e.g., \emph{S. 8 - blur} means strategy 8 and the blurred images are used for testing).}\label{fig:res_comp}
\end{figure*}

As shown in Table~\ref{tab:results}, the results exhibit that training strategy with low quality samples improves the performance on the corresponding distorted images. For instance, the classification accuracy of Strategy 2 only decays about 1\% when the distortion level rises from pristine to level 1. Such decreasing speed is an order of magnitude slower than that of the baseline strategy. However, it is noticed that the performance of these strategies on high-quality pristine images cannot approach as high as the baseline method, which can be intuitively observed in Fig.~\ref{fig:res_comp} (where the dot lines denotes the baseline model while the blue lines denote the models naively trained on mixture data). As discussed in Section~\ref{sec:iqals}, this is due to the fact that 0-1 label distribution teaches the model to be equally confident about the classification results of both high and low quality images.

\subsubsection{Training with IQA-based label smoothing}

Strategies 3,5,7,9 in Table~\ref{tab:results} target at training the model based on the mixture data as well. Moreover, in contrast to the previous strategies, the label distribution is regularized by the proposed IQA-based label smoothing method.

As shown in Fig.~\ref{fig:res_comp}, the performance of models trained with IQA-LS is credibly better than the original ones on relative high quality images (e.g.,pristine, distortion level 1 and 2). Regarding to the performance on strongly distorted images (e.g., distortion level 3), although the models trained with IQA-LS are slightly weaker than those without IQA-LS, the performance drop is marginal and acceptable. Therefore, it is concluded that, comparing to the straightforward way that trains the deep models on mixture data, our proposed IQA-LS technique is not only effective in maintaining the high classification performance for distorted samples, but also promising in improving the accuracy on high quality test data. Moreover, from Fig.~\ref{fig:res_comp}(d), it is observed that when training on mixture of samples with multiple types of artifacts rather than a certain type the superiority of IQA-LS is more apparent. This can be explained by the reason that the regularized label distribution penalises the false inference based on the quality levels, which provides the DNN models with stronger generalizing ability.

\begin{figure}
	\centering
	\includegraphics[width=82mm]{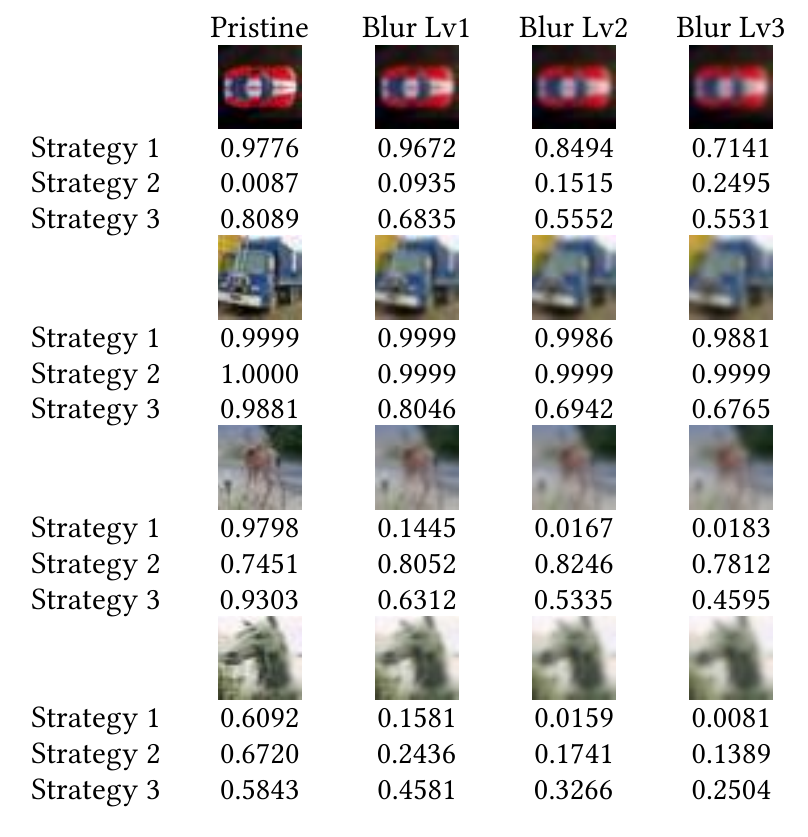}
	\vspace{-3 mm}
	\caption{Examples of the test images and the corresponding confidence values predicted by models trained with Strategy 1 (baseline), Strategy 2 (trained on mixture data straightforwardly) and Strategy 3 (trained on mixture data with proposed IQA based label smoothing technique).}\label{fig:discuss}
\end{figure}

\subsubsection{Discussions}
Here, we perform an in-depth analysis on the proposed scheme to gain a better understanding on the IQA based label-smoothing technique. Let us focus on a certain type of distortion, i.e. blur. Fig.~\ref{fig:discuss} provides some example images along with the confidence values that indicate the prediction results for the correct class with three strategies, which are the baseline and the two trained on $MIX_{blur}$. In particular, the confidence values are generated by the corresponding node of the softmax layer. Specifically, the value close to 0 implies that the model is unconfident about the prediction, while the value close to 1 corresponds to a high confidence prediction. It can be observed that Strategy 3, which utilizes IQA-based label smoothing technique, can always provide a smooth and moderate variation trend on the confidence values with the increase of the blur level. This is in line with the human perception, as blurry images may usually lead to uncertainty in image understanding, thus creating lower confidence values. By contrast, the confidence values of Strategies 1 and 2 are more consistent when the distortion level changes. For some cases, the prediction cannot be robust when the distortion level is extremely high (the images in the third and fourth rows of Fig.~\ref{fig:discuss}). Moreover, Strategy 2 may suffer from the over-fitting problem to the high-level distorted images (images in the first row of Fig.~\ref{fig:discuss}).

Therefore, the proposed IQA-based label smoothing technique leads to the model that not only provides fairly good prediction performance for both high and low quality images, but also simulates human-like perception from the perspective of uncertainty. Moreover, the proposed method may also help to reduce the over-fitting problem when severely distorted images are used for training. In the future, more distortion types, IQA methods and deep learning models will be investigated in this framework.

\section{Conclusion}
\label{sec:conclusion}
We have proposed a quality assessment based label smoothing approach for deep neural network learning. The novelty of the proposed approach lies in that the distorted images are included in the training process in learning the reliable neutral network model, and IQA is adopted in regularizing the label distribution of training samples to obtain a more robust representation. The performance of the proposed scheme is evaluated based on image classification and it is shown that the proposed scheme achieves high prediction accuracy across different distortion types and levels.

\newpage
\bibliographystyle{ACM-Reference-Format}
\bibliography{sample-sigconf_bibtex}

\end{document}